\documentclass[runningheads]{llncs}

\newif\ifECCVReview

\usepackage[mobile]{eccv}

\usepackage{eccvabbrv}

\usepackage{graphicx}
\usepackage{booktabs}
\usepackage{eso-pic}          
\usepackage[accsupp]{axessibility}

\usepackage[pagebackref,breaklinks,colorlinks,citecolor=eccvblue,urlcolor=blue]{hyperref}

\newcommand{\citesp}[1]{\cite{#1}}

\begin{document}

\title{From Events to Enhancement: A Survey on Event-Based Imaging Technologies}
\titlerunning{From Events to Enhancement}

\author{Yunfan Lu\inst{1} \and
Xiaogang Xu\inst{2} \and
Pengteng Li\inst{1} \and
Yusheng Wang\inst{3} \and
Yi Cui\inst{1} \and
Huizai Yao\inst{1} \and
Mingchao Xu \inst{1} \and
Hui Xiong\inst{1}\thanks{Corresponding author.}}
\authorrunning{Y. Lu et al.}

\institute{The Hong Kong University of Science and Technology (Guangzhou) \and
The Chinese University of Hong Kong \and
The University of Tokyo
}

\AddToShipoutPictureFG*{\AtPageUpperLeft{\raisebox{-8mm}{\makebox[\paperwidth][c]{\footnotesize\color{gray}\shortstack[c]{This paper has been accepted for publication at the\\European Conference on Computer Vision (ECCV) Workshop, 2026}}}}}

\maketitle

\begin{abstract}
\noindent Event cameras offering high dynamic range and low latency have emerged as disruptive technologies in imaging.
Despite growing research on leveraging these benefits for different imaging tasks, a comprehensive study of recent advances and challenges is still lacking.
This limits the broader understanding of how to utilize events in universal imaging applications.
In this survey, we first introduce a physical model and the characteristics of different event sensors as the foundation.
Following this, we highlight the advancement and interaction of image/video enhancement tasks with events.
Additionally, we explore advanced tasks, which capture richer light information with events, \eg~light field estimation, multi-view generation, and photometric imaging.
Finally, we discuss new challenges and open questions offering a perspective for this rapidly evolving field.
\ifECCVReview
\else
Continuously updated resources are available at:
{\footnotesize\texttt{\href{https://github.com/yunfanLu/Awesome-Event-Imaging}{github.com/yunfanLu/Awesome-Event-Imaging}}}
\fi
\end{abstract}

\section{Introduction}
Designing robust and efficient machine imaging systems is a fundamental goal in machine imaging technique.
Traditional frame-based cameras, due to their synchronous output nature, have struggled to achieve this goal \citesp{gehrig2024low}.
In contrast, biology-inspired event cameras asynchronously output event streams, offering significant advantages, \eg~ high dynamic range ($>120 dB~ vs.~ 60dB$), low latency ($<0.1 ms ~vs. ~10ms $), low power consumption \citesp{finateu20205} and low bandwidth.
Therefore, event cameras are promising sensing technologies for next-generation imaging systems.
As a result, the event-based vision community has grown rapidly in recent years~\citesp{chakravarthi2024recent}.
Additionally, several surveys showcase the current state of development in event-based vision.
For example, \citesp{gallego2020event} systematically reviewed its various aspects, \citesp{zheng2023deep} focused on its integration with deep learning, and works \citesp{iddrisu2024event,shariff2024event,ghosh2024event} explored advancements in specific applications like eye tracking, autonomous driving, and depth estimation.
\textit{However, the basic imaging technologies with events have not been systematically studied, and a unified understanding of these achievements, challenges, and future directions remains lacking.}

\begin{figure*}
\centering
\includegraphics[width=\linewidth]{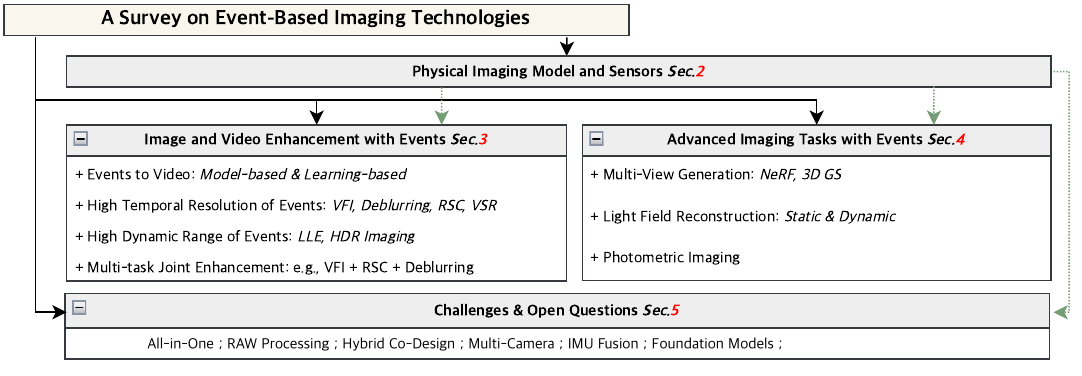}
\caption{A diagram that summarizes this survey.}
\label{fig:3-Overview}
\end{figure*}

Imaging technologies are designed to capture and process analog signals of the sensor, reconstructing light signals of interest, \eg  color and intensity, that serve as inputs for visual perception and understanding tasks.
To systematically and comprehensively study these imaging technologies with events, we first propose a physical imaging model in Sec.~\ref{sec:phy-model}, which describes the process of converting light signals into analog and digital signals, forming the foundation for imaging tasks.
The light signals are described using a plenoptic function~\citesp{Chan2014}, and different imaging technologies aim to recover a portion of this function as the output.
The discussion also covers diverse sensors, highlighting the differences between traditional cameras and event cameras, and illustrating the complementary characteristics and advantages of event cameras.

Building on this physical model, we classify event-based imaging technologies based on the richness of the reconstructed light signals into two primary categories:
\textbf{(I) \textit{enhancement methods for images and videos}}, which are the most commonly used and have received the most attention, and \textbf{(II) \textit{advanced imaging techniques}}, which aim to recover richer light information, \eg  light fields \citesp{habuchi2024time} and photometry \citesp{10655016}.
Enhancement methods for images and videos with events focus on improving visual quality, as discussed in Sec.~\ref{sec:image-video}.
These methods primarily address two research questions: (1) how to leverage high temporal resolution, and (2) how to exploit high dynamic range from events.
For the first question, we specifically explore tasks, \eg  video frame interpolation, deblurring, rolling shutter correction, and video super-resolution.
Regarding the second question, we cover tasks such as low-light image enhancement and HDR imaging.
Moreover, we extend the discussion to more complex, multi-task imaging challenges~\citesp{lu2024uniinr,kim2024frequency}, exploring the relationships between different tasks and potential unified solutions.

Advanced imaging techniques extend beyond traditional images and videos by incorporating richer light information, as discussed in Sec.~\ref{sec:advanced}. These methods not only create new opportunities for multi-view generation and advanced applications, but also introduce significant challenges.
While the information recovered in this group surpasses that of conventional images and videos, many techniques originally designed for video processing remain highly relevant and have been successfully integrated into these advanced methods. Building on this, we identify several promising research directions and propose new themes for future investigation.

Finally, we discuss open questions of event-based imaging technologies in Sec.~\ref{sec:challenges}, which study both established questions in the academic community, \eg ~progress in the RAW domain, and highlight areas that have yet to receive much attention, \eg  the co-design of sensor and algorithms, and other research with practical applications.
In conclusion, this survey provides a comprehensive overview of the current state of event-based imaging, offering valuable insights and directions for future research to drive further advancements in the field.

\section{Physical Imaging Model and Sensors\label{sec:phy-model}}
Imaging models describe the conversion of light into analog and digital signals, forming the physical foundation for imaging tasks, which use digital signals to reconstruct the light signals of interest. Therefore, this section first covers the process of imaging models, and then explores the practical characteristics of sensors, setting the stage for the following sections.

\begin{figure}[t!]
\centering
\includegraphics[width=0.6\linewidth]{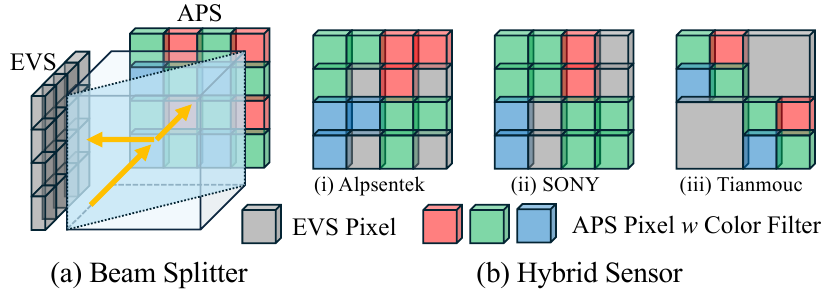}
\caption{Illustration of two imaging systems combining event-based vision sensors (EVS) and active pixel sensors (APS): (a) Beam-splitter-based systems use two separate sensors, with EVS, \eg  PROPHESEE \protect\citesp{finateu20205}; (b) Hybrid sensor, such as (i) Alpsentek \protect\citesp{alpsentek}, (ii) SONY \protect\citesp{kodama20231}, and (iii) Tianmouc \protect\citesp{yang2024vision}.}
\label{fig:4-Sensor}
\end{figure}

\noindent\textbf{Physical Imaging Model:} The plenoptic function \citesp{Chan2014} represents the original light signals.
Specifically, the plenoptic function defines the intensity of every light ray in a scene, as represented by Eq.~\ref{eq:plenoptic_function}, where \(x, y, z\) are spatial coordinates relative to the camera optical center \(O\), \( \theta, \phi \) are the elevation and azimuth angles, \( \lambda \) is the wavelength, and \( \tau \) represents time.
This plenoptic function thus offers a complete representation of light signals, and imaging tasks generally recover only a subset of this information.
\begin{equation}
L(x, y, z, \theta, \phi, \lambda, \tau) \label{eq:plenoptic_function}
\end{equation}
Considering a camera sensor composed of \(N\) rows and \(M\) columns of pixels, each pixel can be treated as recording the intensity of a light ray that arrives at it.
Therefore, placing the sensor plane at \(z = f\) in the camera coordinate system, where \(f\) is the focal length, the position of each pixel \((i, j)\) on the sensor plane is described by Eq.~\ref{eq:pixel_position}.
Here, \(c_x, c_y\) denote the pixel coordinates of the sensor's center (\(M/2, N/2\)), \(d\) represents the physical pixel spacing, \(u=(j-c_x)d\), and \(v=(i-c_y)d\).
\begin{equation}
((j - c_x) \cdot d, (i - c_y) \cdot d, f)
\label{eq:pixel_position}
\end{equation}
Thus, the direction of the light ray corresponding to the pixel \((i, j)\) is determined by the vector from the optical center \(O = (0, 0, 0)\), that passing through \((u, v, f)\) on the sensor plane.
The calculations for the elevation angle is \(\theta_{ij} = \arccos\left( f / \sqrt{u^2 + v^2 + f^2} \right)\) and azimuth angle is \(\phi_{ij} = \arctan2(v, u)\).
Consequently, the light ray associated with pixel \((i, j)\) can be described using the plenoptic function \(L(x, y, z, \theta, \phi, \lambda, \tau)\), as shown in Eq.~\ref{eq:camerap_p}.
Here, \(R(i, j, \lambda)\) represents the camera's response function, indicating the sensor's sensitivity to different wavelengths, which is typically determined by filters such as RGB filters.
\begin{equation}
P(i, j, \tau) = \int_{\lambda} L\left(O, \theta_{ij}, \phi_{ij}, \lambda, \tau\right) \cdot R(i,j,\lambda) \, d\lambda
\label{eq:camerap_p}
\end{equation}
In the sensor, the light signal \(P\) is converted into an analog signal $A$ through photoelectric conversion $f_{pc}$, which introduces Gaussian noise $N_{GS}$ and Poisson noise $N_P$, both of which are also influenced by the $P$, as simplified in Eq.~\ref{eq:light_to_electrical}.
\begin{equation}
A(i, j, \tau) = f_{pc}(P(i, j, \tau)) + N_{GS} + N_P
\label{eq:light_to_electrical}
\end{equation}
For frame-based cameras, the final RGB frame is obtained after exposure and signal processing \(f_{\text{i}}\).
The frame $I_0$ at time \(t_0\) is expressed as shown in Eq.~\ref{eq:isp}.
\begin{equation}
I_{0} = f_{i}( \int_{t_0}^{t_0+\Delta t} A(t) \, dt)
\label{eq:isp}
\end{equation}
The output of event cameras is based on changes in voltage within the logarithmic domain.
The cumulative change is denoted as \( \Delta\hat{A} = \log A(i, j, t_0) - \log A(i, j, t') \), where \( t' \) is the last trigger time.
The threshold \( \theta_{\text{th}} \) controls the sensitivity.
The event \( E(i, j, t_0) \) is determined as Eq.~\ref{eq:event_generate}.
\begin{equation} \label{eq:event_generate}
E(i, j, t_0) =
\begin{cases}
+1 & \Delta\hat{A} > ~~~\theta_{\text{th}} \\
-1 & \Delta\hat{A} < -\theta_{\text{th}} \\
0 & \text{otherwise}
\end{cases}
\end{equation}
\noindent\textbf{Sensors:}
In practice, event-camera-based imaging systems can be broadly categorized into three types: (1) systems using only event cameras, (2) beam-splitter-based systems, and (3) hybrid sensor-based systems.
Systems (2) and (3), shown in Fig.~\ref{fig:4-Sensor}, have gained more attention as they capture both the integral, Eq.~\ref{eq:isp} and differential, Eq.~\ref{eq:event_generate}, components of light signals simultaneously.
Beam-splitter-based systems \citesp{tulyakov2021time,wang2023event} use separate EVS and APS chips, which are easier to acquire.
However, the optical prisms increase the size of the imaging device and introduce challenges related to the alignment of the two sensors.
In contrast, hybrid sensors \citesp{alpsentek,kodama20231,yang2024vision} integrate EVS and APS at the chip level, ensuring temporally and spatially aligned data streams.
However, such systems are limited by the scarcity of publicly available datasets at this time and challenges associated with stacked fabrication processes, which introduce noise and other issues \citesp{wu2024mipi}.

\section{Image and Video Enhancement with Events\label{sec:image-video}}
The generation of images and video frames is based on Eq.~\ref{eq:isp}. However, due to the limitation of only recording the integral part of the analog signal in Eq. \ref{eq:isp}, several degradations arise, such as insufficient frame rates and motion blur. Thanks to the additional differential information provided by event cameras, as shown in Eq. \ref{eq:event_generate}, enhancing images and videos with events is one of the most prominent technical tasks in event-based imaging.
These methods can be categorized into four groups:
\textbf{(1)} events to video, Sec. \ref{sec:event_to_video};
\textbf{(2)} leveraging the high temporal resolution of events, Sec. \ref{sec:high_temporal_resolution};
\textbf{(3)} utilizing the high dynamic range of events, Sec. \ref{sec:high_dynamic_range};
\textbf{(4)} multi-task learning with events, Sec. \ref{sec:multi-task}.

\subsection{Events to Video\label{sec:event_to_video}\label{Ses:E2VID}}
\textbf{\textit{Overview}:}
Event-to-video methods involve reconstructing videos solely from event streams,
essentially reconstructing Eq.~\ref{eq:isp} based on Eq.~\ref{eq:event_generate},
which directly leverage the high dynamic range and high temporal resolution of events.
We categorize event-to-video methods into two main types: \textbf{(1)} model-based and \textbf{(2)} learning-based methods, depending on the underlying technique used for reconstruction.

\noindent\textbf{(1) Model-Based Methods:}
These methods establish explicit rules to directly map events to videos.
Early approaches relied on hand-crafted rules.
For example, \citesp{bardow2016simultaneous,duwek2021image} explored the relationship between image gradients and optical flow using photometric constancy, employing optical flow to reconstruct videos.
Building on this, \citesp{zhang2022formulating} framed optical flow and image brightness as a linear inverse problem to enhance the robustness.
More recently, \citesp{wang2025revisit} introduced an implicit neural representation that directly reconstructs intensity from events through the event-generation partial differential equation, without requiring optical-flow estimation.
\textbf{\textit{Key Insights:}}
These methods excel in leveraging rules for scene-specific reconstruction, providing strong interpretability and generalization to new cameras and scenes.
However, because these methods rely only on events from the current scene, they may lose information in regions with sparse events and cannot exploit learned priors.

\noindent\textbf{(2) Learning-Based Methods}:
These methods train models on large datasets and apply the trained models to new scenes.
Learning-based methods can be categorized into two types: supervised and unsupervised approaches.
Supervised methods typically involve learning a temporal network, \eg  RNN, that transforms the event stream into intensity for video frames.
Representative examples include E2VID~\citesp{rebecq2019high}, and its extensions~\citesp{cadena2021spade,ercan2024hypere2vid}.
These methods establish connections between current events and prior frames to predict the current frame.
However, they rely on frames for supervision, which are typically captured with low dynamic range and low temporal resolution, limiting the ability to fully leverage the advantages of event cameras.
In contrast, unsupervised methods, such as \citesp{paredes2021back}, aim to directly establish relationships between optical flow, images, and events within the event stream.
These methods use self-supervised learning to reconstruct videos without relying on ground truth frames, thus pushing the model's performance limits.
However, both supervised and unsupervised methods face challenges due to the inherent sparsity and differential nature of events, which make the reconstruction results ill-posed.
A recent work, LaSe-E2V~\citesp{chen2024lase}, introduces diffusion models with language guidance \citesp{rombach2022high}, incorporating semantic information into the reconstruction process.
While this helps alleviate the ill-posed nature of the task, it introduces a new limitation: the semantic information is generated by models rather than captured by real cameras.
\textbf{\textit{Key Insights:}} Event-to-video methods are limited by the ill-posed nature of the problem, challenges in dataset collection, and the lack of semantic information.
As a result, increasing attention has been focused on integrating events with RGB images to address these limitations.

\begin{figure}[t!]
\centering
\includegraphics[width=0.9\linewidth]{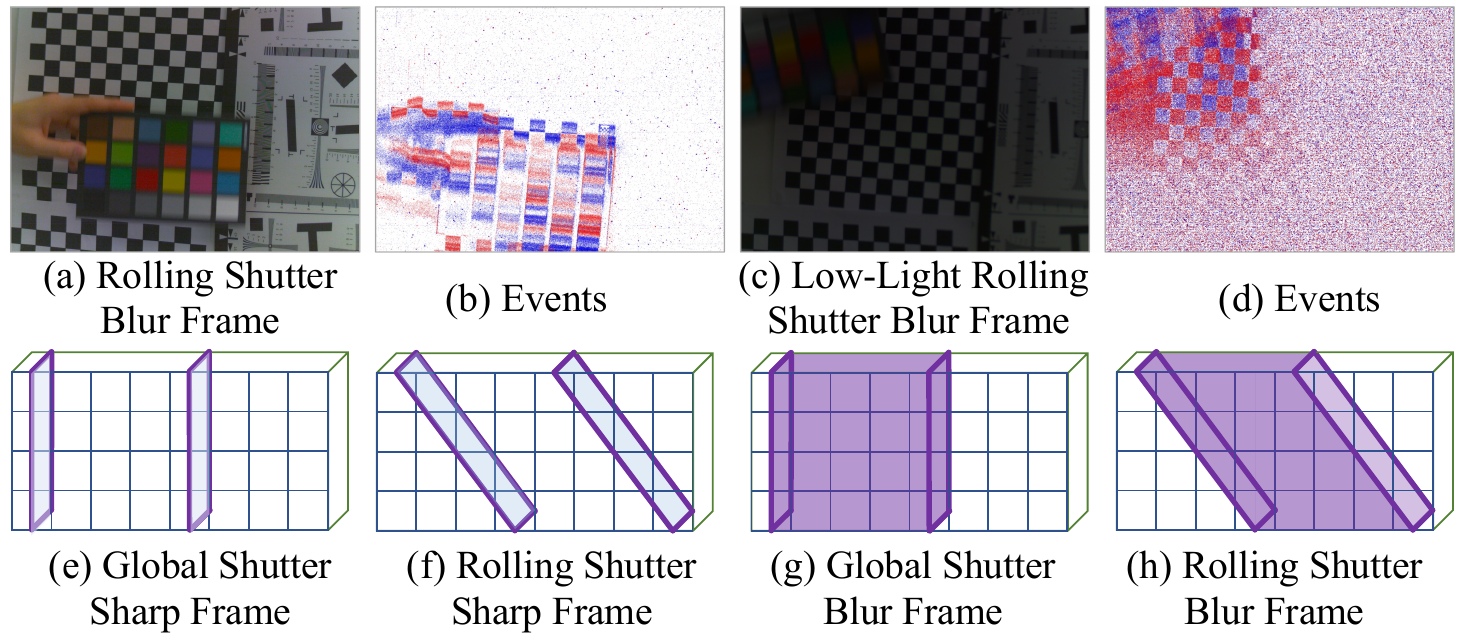}
\caption{(a-d) show the RGB frames and events in real scenes. (e-h) show diagrams of different degradations.}
\label{fig:1-Sample-Release}
\end{figure}

\subsection{High Temporal Resolution of Events\label{sec:high_temporal_resolution}}
\textbf{\textit{Overview}:}
In this section, we examine how the high temporal resolution of events can be exploited for video enhancement.

Unlike traditional cameras—which suffer from temporal degradations such as low frame rates, motion blur, and rolling shutter artifacts—event cameras capture fine-grained motion information with high temporal resolution.
Consequently, researchers focus on harnessing this temporal fidelity to improve motion representation in video generation.

\noindent\textbf{Video Frame Interpolation (VFI) } aims to transform low-frame-rate videos into high-frame-rate ones.
Traditional interpolation methods are limited by strict assumptions about illumination and motion, which prevent them from modeling complex nonlinear motion. Event-based methods have addressed these limitations. For instance, TimeLens~\citesp{tulyakov2021time} initiated this line of research by proposing a dual-stream approach. One stream synthesizes frames from events, while the other estimates optical flow using events to warp adjacent frames. The result is obtained by combining the outputs of both streams. Similarly, TimeReplayer~\citesp{he2022timereplayer} extends this methodology in a self-supervised manner.
However, due to the inherently sparse nature of event triggers, illumination changes and sensor noise can also trigger events, so not all events correspond to scene motion.
To address these challenges, TimeLens++ \citesp{tulyakov2022time} introduces approximation methods that parameterize complex motion between frames. Additionally, CBMNet \citesp{kim2023event} proposes a hybrid approach that combines dense visual information from images with bidirectional sparse motion information from events, while accounting for the modality differences between these two data types.
More recently, TimeLens XL~\citesp{ma2025timelens} has been proposed, which decomposes large inter-frame motion into smaller motion segments and recursively reconstructs the full motion trajectory.
\textbf{\textit{Key Insights:}} Event-based VFI has made significant progress in robust motion estimation under illumination changes, modeling complex nonlinear motion, and handling large displacements.
Nonetheless, a major limitation of these methods is their heavy reliance on datasets acquired using beam-splitter systems, which may hinder their applicability in real-world applications.

\noindent\textbf{Deblurring} focuses on modeling intra-frame motion information to leverage the high temporal resolution of event cameras, as shown in Fig.~\ref{fig:1-Sample-Release}.
Early works primarily modeled the relationship between blurry frames (Eq.~\ref{eq:isp}) and events (Eq.~\ref{eq:event_generate}),
\citesp{EDI} proposed an optimization-based method and \citesp{xu2021motion} proposed self-supervised frameworks.
Subsequent studies identified limitations in consecutive blurry frame assumption: \citesp{shang2021bringing} challenged it by introducing nearest sharp frame detection, while \citesp{sun2022event} constructed a real-world dataset and employed cross-modal attention to filter noisy events and model realistic blur.
More recently, to handle real-world challenges, \citesp{kim2022event} addressed dynamic exposure time estimation, and \citesp{zhang2023generalizing} proposed scale-aware networks for varying spatial-temporal blur distributions.
Hardware constraints were also considered—\citesp{cho2023non} tackled non-coaxial alignment between event and RGB cameras, while \citesp{ma2024color4e} resolved color restoration through event demosaicing and bimodal fusion.
Recent advancements by \citesp{kim2024frequency} leveraged long-range temporal dependencies using frequency-aware features and introduced hybrid-camera real-world datasets.
\textbf{\textit{Key Insights:}}
Progress has been achieved through physical modeling, adaptive feature fusion, real-world generalization, and hardware-aware optimization.
The practical explorations in these works also provide valuable insights for other event-based imaging tasks.

\noindent\textbf{Rolling Shutter Correction (RSC)} aims to eliminate distortions caused by inconsistent exposure times in RS images, especially in fast-moving scenes, as shown in Fig.~\ref{fig:1-Sample-Release}. Unlike motion blur, RS artifacts require precise intra-frame motion estimation due to row-wise exposure differences. We classify existing methods into two groups: (1) Supervised and (2) Self-supervised methods.
\noindent\textit{(1) Supervised:}
EvUnRoll \citesp{zhou2022evunroll} introduced a dual-stream network—one branch corrects edge distortions using optical flow, while the other recovers occluded regions—assuming constant motion. To accommodate variable motion, EvShutter \citesp{erbach2023evshutter} added an event transformation module that encodes the differences between global shutter (GS) and RS images. Further, \citesp{wang2024event} employed a time-guided attention mechanism to reduce temporal ambiguities and enhance motion estimation.
\noindent\textit{(2) Self-supervised:}
Due to challenges in acquiring aligned RS, GS, and event data, self-supervised methods \citesp{wang2023self,lu2023self} explore bidirectional transformations (RS-to-GS and GS-to-RS) to reconstruct GS images without direct supervision. Notably, \citesp{lin2023event} introduced a quad-camera system capable of capturing aligned RS, GS, and event datasets, offering a valuable resource for validation.
\textbf{\textit{Key Insights:}}
These methods show progress in motion modeling and artifact correction, but their evaluation on simulated or limited datasets raises concerns about real-world applicability. Addressing domain gaps and integrating multi-task learning to jointly correct RS artifacts and related degradations is a promising future direction.

\noindent\textbf{Video Super Resolution (VSR)} aims to enhance the spatial resolution of videos. By leveraging events, which provide high-frequency temporal information, these methods convert temporal details into high-frequency spatial features. \citesp{jing2021turning} first proposed a framework that performs video frame interpolation followed by multi-frame fusion for VSR. However, the interpolation and fusion stages introduce cumulative errors. 
To mitigate these issues, \citesp{lu2023learning} combines a Spatial-Temporal Fusion (STF) module, a Temporal Filter (TF) module, and a Spatial-Temporal Implicit Representation (STIR) module to exploit the complementary spatiotemporal information from frames and events for reconstruction.
Similarly, \citesp{kai2023video} employed optical flow for temporal alignment, thereby enhancing multi-frame fusion and VSR outcomes. In contrast, \citesp{kai2024evtexture} prioritized learning texture information from events rather than explicitly modeling spatiotemporal motion. Furthermore, \citesp{xiao2025event} proposed a lightweight adapter that enables traditional VSR methods to benefit from events by aligning multi-frame inputs with motion cues derived from events, achieving superior results without major modifications to existing frameworks.
\textbf{\textit{Key Insights:}}
Event-based VSR methods have evolved from direct interpolation and fusion toward advanced strategies that effectively capture both motion and texture features. Nonetheless, cumulative errors and optimal integration of events remain open challenges.

\noindent\textit{\textbf{Summary:}} The high temporal resolution of events enables precise modeling of diverse motions, including inter-frame (VFI), intra-frame (deblurring), and inter-row (rolling shutter correction), effectively addressing key limitations of conventional imaging. Additionally, integrating motion information to enhance spatial details for super-resolution remains challenging. Moving forward, a unified framework for multi-task learning and modeling is essential.

\subsection{High Dynamic Range of Events\label{sec:high_dynamic_range}}
\textbf{\textit{Overview:}}
The high dynamic range of events provides significant advantages in low-light conditions and for expanding overall scene dynamic range. However, effectively integrating event data with images remains challenging due to the modality gap and the difficulty of constructing aligned datasets, which are key research challenges in this area.

\noindent\textbf{Low-Light Enhancement (LLE)} aims to restore images captured in dim conditions to normal lighting levels. Traditional cameras struggle with severe noise and loss of edge and texture details in dark regions, whereas events inherently provide a broader dynamic range. Early methods primarily focused on the network architecture design. For example, \citesp{zhang2020learning} pioneered the use of adversarial networks to bridge the modality gap between events and images. Building on this, \citesp{jiang2023event} introduced a residual fusion module and a multi-scale mechanism to further mitigate modality differences and enhance perceptual quality. To extend single-frame enhancement to videos, \citesp{liang2023coherent} proposed a spatiotemporal network that jointly models spatial and temporal correlations between events and video frames, effectively transferring HDR properties from events to low-light video sequences. Although these methods have shown promising results, they remain insufficiently validated on real-world datasets. A major challenge is the scarcity of large-scale, aligned datasets—an issue partially addressed by a recent work \citesp{liang2024towards}, which developed an optical system using a robotic arm and ND filters to capture synchronized normal-light and low-light data.
\textbf{\textit{Key Insights:}} While event-based low-light enhancement has made significant progress in both methods and datasets, the lack of a unified standard for defining normal-light images in existing datasets limits network performance.

\noindent\textbf{HDR Imaging} seeks to enhance the overall dynamic range of images, representing a more ambitious goal than low-light enhancement. Early work \citesp{han2023hybrid} fused LDR images with intensity reconstructions from events to generate HDR images, but cumulative reconstruction errors degraded the dynamic range. In contrast, \citesp{messikommer2022multi} used events as a source of motion information to align multi-exposure images for fusion, yet this approach did not fully exploit the HDR potential of events. To address these limitations, \citesp{yang2023learning} introduced a shared latent space to align events with LDR frames, capturing complementary dynamic range information while enforcing temporal coherence through an event-modulated consistency mechanism. A recent advancement by \citesp{cui2024color} incorporates color events via a color-event HDR dataset, directly leveraging the HDR inherent in events.
\textbf{\textit{Key Insights:}} HDR imaging with events remains hampered by reconstruction errors due to the lack of high-quality ground truth for supervision.

\noindent\textbf{\textit{Summary:}}
The HDR of events drives advancements in both low-light enhancement and HDR imaging. However, the lack of high-quality datasets and a well-defined task formulation remain major barriers to progress in this field. Future work should focus on developing comprehensive and robust datasets to fully unlock the HDR potential of event sensors.

\subsection{Multi-task Joint Enhancement \label{sec:multi-task}}
\textbf{\textit{Overview:}}
In practical imaging scenarios, multiple degradations—such as blur, low-light, and rolling shutter distortions—often occur simultaneously. Addressing these intertwined degradations is challenging yet highly valuable. In this section, we review methods that integrate multiple tasks by leveraging the high temporal resolution and dynamic range of event cameras. Specifically, we categorize these methods into two groups: (1) tasks addressing temporal degradation and (2) tasks combining temporal degradation with dynamic range enhancement.

\noindent\textbf{Multi-task Integration for Temporal Degradation:}
Recent works exploit event cameras to mitigate intertwined temporal degradations by capitalizing on their motion-aware properties. For example, in \textit{VFI + Deblurring} tasks, \citesp{yang2024latency} introduces latency correction to reduce misalignment between events and blurry frames, while \citesp{weng2023event} employs event-guided mutual constraints to tackle blind exposure estimation. Moreover, \citesp{sun2023event} generalizes this approach with a bidirectional recurrent network that adaptively fuses events and frames for joint deblurring and interpolation, as validated on the HighREV dataset. In addition, \citesp{song2022cir} reconstructs continuous sharp videos from blurry inputs using event-guided parametric functions, and \citesp{zhang2022unifying} unifies deblurring and interpolation within a self-supervised framework that exploits mutual constraints among events, blurry frames, and latent sharp images. Furthermore, for \textit{VFI + Super Resolution}, \citesp{lu2024hr} proposes an implicit neural representation (INR) framework supported by event temporal pyramids to enhance spatiotemporal resolution. More complex scenarios, such as \textit{VFI + Deblurring + Rolling Shutter Correction}, are addressed by \citesp{zhang2024neural}, which introduces a unified neural re-exposure framework with cross-modal attention, and by \citesp{lu2024uniinr}, which employs spatial-temporal INR to directly map degraded inputs to high-frame-rate, sharp outputs.
\textbf{\textit{Key Insights:}}
These studies suggest that a unified temporal enhancement framework may be feasible; however, designing effective spatiotemporal representations remains a challenge.

\noindent\textbf{Dynamic Range Enhancement Combined with Temporal Degradation:}
Event cameras also enable the joint enhancement of dynamic range and temporal fidelity. For \textit{Deblurring + LLE}, \citesp{kim2024towards} introduces a hybrid camera system with an end-to-end framework that suppresses noise while recovering sharp details in low-light videos.
In the case of \textit{VFI + LLE}, \citesp{zhang2024sim} uses per-scene optimization to mitigate trailing artifacts under low-light conditions and introduces the EVFI-LL dataset for evaluating generalization across different lighting conditions.
\textbf{\textit{Key Insights:}}
Although jointly leveraging high dynamic range and temporal resolution is promising, exploration in this direction is still limited and requires further breakthroughs.

\noindent\textbf{\textit{Summary:}}
Collectively, these works mark a paradigm shift from task-specific methods to unified frameworks. Event cameras serve as a cross-modal bridge to address intertwined spatial, temporal, and photometric degradations via adaptive fusion, implicit neural representations, and self-supervision. This progression highlights the potential of event-based systems to manage the real-world complexity of multi-task imaging scenarios.

\section{Advanced Imaging Tasks with Events\label{sec:advanced}}
Advanced imaging tasks go beyond the limitations of image and video, aiming to recover richer optical information. While traditional RGB-based techniques have been explored in these areas extensively, event-based approaches remain relatively underdeveloped. Current works primarily focus on three key aspects: multi-view generation, Sec. \ref{sec:nerf-3dgs}; light field reconstruction Sec. \ref{sec:light-field}, and photometric imaging, Sec. \ref{sec:photometric}.

\subsection{Multi-View Generation\label{sec:nerf-3dgs}}
Multi-View Generation aims to reconstruct high-dimensional 3D scene representations from given camera coordinates, enabling the synthesis of novel views, restoring camera pose flexibility as described in Eq.~\ref{eq:plenoptic_function}.
Unlike conventional image and video tasks that capture two-dimensional information, multi-view generation provides richer spatial details.
Event cameras excel in this domain due to their robustness under challenging conditions (\eg~ motion blur and low light), making them a valuable complement to traditional imaging.

\noindent\textbf{Neural Radiance Fields (NeRF)-based Methods:}
Early event-based approaches, such as EventNeRF~\citesp{EventNerf} and E-NeRF~\citesp{E-NeRF}, directly optimize neural radiance fields from events captured by a moving event camera. However, these methods often assume ideal event imaging conditions. Robust $e$-NeRF~\citesp{Robust-e-NeRF} therefore models intrinsic parameters such as contrast thresholds and refractory periods. E$^3$NeRF~\citesp{E3NeRF} exploits spatiotemporal event cues and spatial attention to recover texture from blurry images, whereas Deblur $e$-NeRF~\citesp{Deblur-e-NeRF} uses a physically accurate pixel-bandwidth model together with a threshold-normalized total-variation loss.
\textbf{\textit{Key Insights:}} NeRF-based methods demonstrate promise in leveraging event data for 3D reconstruction while remaining constrained by sensitivity to noise in low-quality event streams.

\noindent\textbf{3D Gaussian Splatting (GS)-based Methods:}
3D GS-based methods blend continuous Gaussian primitives to render scenes efficiently, but mapping sparse and noisy intensity changes to geometry, appearance, and camera motion remains difficult.
For instance, Event3DGS\citesp{Event3DGS} identifies a \textit{neutralization phenomenon} where accumulated positive and negative events degrade reconstruction quality, addressing this with noise models that better reflect event sparsity and noise. BeSplat~\citesp{BeSplat} jointly reconstructs 3D Gaussians and Bezier $\mathrm{SE}(3)$ camera motion from a single blurry image and its event stream, whereas Ev-GS~\citesp{Ev-GS} reconstructs a 3D Gaussian representation from a monocular event camera using pure event-based supervision.
Meanwhile, SweepEvGS~\citesp{SweepEvGS} explores efficient sweeping strategies to fuse multi-view event data under non-uniform camera motion.
EventSplat~\citesp{EventSplat} uses an event-to-video prior for SfM initialization and spline-based pose optimization, enabling real-time rendering but inheriting errors from reconstructed intensity frames. 
IncEventGS~\citesp{IncEventGS} instead jointly refines a continuous camera trajectory and 3D Gaussians from a single event camera without known poses. DiET-GS~\citesp{Lee_2025_CVPR} fuses blurry multi-view RGB, event double-integral constraints, and a pretrained diffusion prior; it recovers richer color and detail at the cost of multimodal input. 
EF-3DGS~\citesp{EF-3DGS} calibrates camera poses by maximizing the contrast of warped events and introduces gradient-domain constraints, while EvGGS~\citesp{EvGGS} incorporates a depth-estimation module as an additional geometric cue.
\textbf{\textit{Key Insights:}}
3D GS improves efficiency, while the 2025 works clarify a central trade-off: event-only, pose-free capture is flexible but underconstrained, whereas image or generative priors improve appearance at the risk of prior-induced artifacts.

\noindent\textbf{\textit{Summary:}} 
Event-based multi-view generation leverages the high temporal resolution and motion sensitivity of event to synthesize novel views in complex imaging conditions such as blur or low light, with NeRF-based and 3D GS-based methods representing two prominent paradigms. While both approaches advance the state of the art, exploring novel representations and integrating realistic camera models remains an important research direction.

\subsection{Light Field Reconstruction\label{sec:light-field}}

\noindent\textbf{\textit{Overview:}}
Light field reconstruction aims to recover a 4D representation.
Denoted by $L(u,v,x,y)$, it is a reduced 4D parameterization of the plenoptic function in Eq.~\ref{eq:plenoptic_function} that captures spatial and angular scene light information for applications like view synthesis and refocusing. Event cameras excel in this task by detecting subtle intensity changes and rapid motion missed by conventional sensors.
Existing event-based light field reconstruction methods can be classified into two categories based on scene dynamics: \textbf{(1)} {Static Scene} and \textbf{(2)} {Dynamic Scene}.

\noindent\textbf{Static Scene:}
\citesp{habuchi2024time} integrates event cameras with coded apertures to capture intensity variations, reconstructing multiple coded-aperture images from a single exposure. Those images train the model for synthesizing the 4D light field.
\textbf{\textit{Key Insights:}} The static scene method enables light field reconstruction in a single exposure but is restricted to static scenarios and specific hardware. Besides, it only relies on stacked event representations, overlooking the potential of continuous event streams.

\noindent\textbf{Dynamic Scene:}
EventLFM \citesp{guo2024eventlfm} integrates event cameras with Fourier light field microscopy for 1-kHz 3D volume reconstruction. For macroscopic scenes, \citesp{qu2024event} uses kaleidoscope- and galvanometer-based multiplexing, with the latter preserving spatial resolution during dynamic capture. S2D-LFE~\citesp{Liu_2025_CVPR} shifts from direct capture to diffusion-based angular densification: an LFE-specific VAE and adapter generate up to $9{\times}9$ views from $2{\times}2$ inputs while maintaining angular-temporal consistency.
\textbf{\textit{Key Insights:}} Dynamic methods improve capture speed or angular density, but optical multiplexing sacrifices hardware simplicity. S2D-LFE avoids dependence on the frame modality, yet its 162-sequence training corpus is simulated; its performance under broader real-world conditions remains to be further evaluated.

\noindent\textbf{\textit{Summary:}}
Static methods perform well in controlled settings, while dynamic methods trade hardware complexity, spatial resolution, and learned priors for temporal or angular density. Hardware simplification and stronger real-data validation remain priorities.

\subsection{Photometric Imaging\label{sec:photometric}}

\textbf{\textit{Overview:}} Photometric imaging estimates object illumination and reflectance.
Traditional methods often struggle with intensity–distance ambiguity and limited dynamic range.
In contrast, events provide a promising alternative in photometric stereo and indoor lighting estimation.

Photometric Stereo estimates surface normals from responses to changing illumination. EFPS-Net~\citesp{Ryoo_2023} fuses RGB and events at object boundaries, while EventPS~\citesp{10655016} estimates normals in real time from events alone. Two CVPR 2025 studies extend this line in complementary ways. PS-EIP~\citesp{Kitazawa_2025_CVPR} fits inverse inter-event-interval profiles and rejects shadow, specular, and noise outliers without learning; on 3D-printed objects, its mean angular error is 8.12 degrees versus 13.66 degrees for EventPS-FCN. EventPSR~\citesp{Yu_EventPSR_2025_CVPR} jointly estimates normals and metallic/roughness reflectance at about 30\% of the frame-based data rate. The former prioritizes robustness and interpretability, whereas the latter recovers richer material properties. Event-based indoor lighting estimation~\citesp{9578832} recovers radiance from high-frequency variations but remains vulnerable to abrupt luminance transients.
\noindent\textbf{\textit{Key Insights:}}
Event-only photometry is becoming more robust and expressive, yet cast shadows, inter-reflections, complex non-Lambertian transport, and calibration remain important challenges.

\section{Challenges \& Open Questions\label{sec:challenges}}
Despite significant progress in event-based imaging, the transition from isolated tasks to robust, integrated systems remains hindered by critical, multifaceted challenges—such as sensor noise, light flicker, and the integration of heterogeneous data.
Furthermore, advancing event-based imaging requires the development of unified and multi-task frameworks and adaptive hardware solutions that bridge the gap between controlled laboratory environments and the complexities of real-world conditions.
The following discussion outlines key challenges and open questions, toward more robust and comprehensive event-based imaging systems.

\noindent\textbf{(1) All-in-One Imaging:}
In real-world applications, event camera systems must address multiple degradation factors simultaneously.
For example, a drone's camera often encounters low-light conditions, motion blur, and rolling shutter distortions all at once.
Although some studies have combined tasks in Sec.~\ref{sec:multi-task}, these approaches remain fragmented.
A deeper exploration of the intrinsic relationships among these challenges and the development of unified, end-to-end frameworks are essential for improving efficiency and robustness through integrated solutions.

\noindent\textbf{(2) RAW Domain Processing with Events:}
As discussed in Sec.~\ref{sec:image-video} and Sec.~\ref{sec:advanced}, most imaging systems assume high-quality events and RGB frames capture.
However, real-world hybrid systems often suffer from missing values and noise in RAW data due to event-generating pixels~\citesp{wu2024mipi}, posing challenges for the image signal processing (ISP) pipeline.
Meanwhile, events offer rich dynamic range and motion cues, presenting an opportunity to enhance ISP processes.
Despite its potential, research in this area remains limited, leaving significant room for exploration.

\noindent\textbf{(3) Hybrid Sensor and Algorithm Co-Design:}
Hybrid sensors that integrate APS and EVS technologies remove complex optical paths and provide spatially aligned frame and event measurements. However, their heterogeneous pixel architectures can introduce missing values, noise, and other RAW-domain artifacts, posing new challenges for image reconstruction and ISP. These characteristics motivate the joint design of sensor architectures and imaging algorithms to better exploit the complementary information from APS and EVS measurements.

\noindent\textbf{(4) Multi-Camera Integration:}
Traditional and event cameras are more accessible than custom hybrid sensors, but cross-camera geometry and timing replace pixel-alignment problems. Active event stereo~\citesp{Li_2025_CVPR} combines two event cameras with projected infrared patterns to recover high-speed depth in dark, textureless, or static regions. Its 21.5k real and 23.8k synthetic labels support inference up to 150 FPS, demonstrating the value of active cues while introducing specialized hardware and calibration. NeRF and 3D GS (Sec.~\ref{sec:nerf-3dgs}) may further relax rigid alignment.

\noindent\textbf{(5) IMU-Guided Imaging:}
IMU inertial sensors have been highly successful in SLAM when combined with event cameras, and this success is promising for new imaging applications as well. IMUs provide additional motion information, making their integration a key challenge in this direction.

\noindent\textbf{(6) Foundation Models with Events:}
Foundation models~\citesp{rombach2022high} offer strong learned priors, but their integration with event-based imaging remains relatively underexplored. A related direction is the use of generative priors. For example, EGDeblurring~\citesp{Xie_2025_CVPR} uses a diffusion model to generate motion-prior guidance from a single blurry image rather than raw event representations. Such approaches highlight the potential of learned generative priors, while motivating further investigation into how these priors can be grounded by real event measurements.

\section{Conclusion\label{sec:conclusion}}
This survey provides a comprehensive overview of event-based imaging by detailing its core principles, diverse applications, and state-of-the-art reconstruction methods. We highlight how event sensors overcome traditional imaging limitations and identify critical challenges that need to be addressed for robust real-world systems.
We hope this study inspires future research and drives the development of more resilient, integrated imaging solutions.

\clearpage
\bibliographystyle{splncs04}
\bibliography{reference}

\end{document}